\documentclass[letterpaper]{article} 
\usepackage{aaai25}  
\usepackage{times}  
\usepackage{helvet}  
\usepackage{courier}  
\usepackage[hyphens]{url}  
\usepackage{graphicx} 
\usepackage{natbib}  
\usepackage{caption} 
\usepackage{algorithm}
\usepackage{algorithmic}

\usepackage{newfloat}
\usepackage{listings}
\DeclareCaptionStyle{ruled}{labelfont=normalfont,labelsep=colon,strut=off} 
\floatstyle{ruled}
\newfloat{listing}{tb}{lst}{}
\floatname{listing}{Listing}
\usepackage{amsmath}
\usepackage{booktabs}
\usepackage{color} 
\usepackage{bm}
\usepackage{soul}

\definecolor{lightgreen}{rgb}{0.7,1,0.7}
\definecolor{lightyellow}{rgb}{1, 1, 0.7}
\definecolor{lightblue}{rgb}{0.7,0.7,1}
\definecolor{lightgray}{rgb}{0.9,0.9,0.9}

\title{Knowledge Editing with Dynamic Knowledge Graphs\\ for Multi-Hop Question Answering}
\author{
    Yifan Lu\textsuperscript{\rm 1},
    Yigeng Zhou\textsuperscript{\rm 1},
    Jing Li\textsuperscript{\rm 1}\thanks{Corresponding author.},
    Yequan Wang\textsuperscript{\rm 2},\\
    Xuebo Liu\textsuperscript{\rm 1},
    Daojing He\textsuperscript{\rm 1},
    Fangming Liu\textsuperscript{\rm 3},
    Min Zhang\textsuperscript{\rm 1}\\
}
\affiliations{
    \textsuperscript{\rm 1}Harbin Institute of Technology, Shenzhen, China\\
    \textsuperscript{\rm 2}Beijing Academy of Artificial Intelligence, Beijing, China\\
    \textsuperscript{\rm 3}Pengcheng Laboratory, Shenzhen, China\\

    lu.yifan@foxmail.com,
    jingli.phd@hotmail.com
}

\usepackage{xspace}
\newcommand{\ourapproach}{\textsc{KeDkg}\xspace}
\begin{document}

\maketitle

\begin{abstract}

	Multi-hop question answering (MHQA) poses a significant challenge for large language models (LLMs) due to the extensive knowledge demands involved.
	Knowledge editing, which aims to precisely modify the LLMs to incorporate specific knowledge without negatively impacting other unrelated knowledge, offers a potential solution for addressing MHQA challenges with LLMs. 
	However, current solutions struggle to effectively resolve issues of knowledge conflicts.
	Most parameter-preserving editing methods are hindered by inaccurate retrieval and overlook secondary editing issues, which can introduce noise into the reasoning process of LLMs. 
	In this paper, we introduce \ourapproach, a novel knowledge editing method that leverages a dynamic knowledge graph for MHQA, designed to ensure the reliability of answers.
	\ourapproach involves two primary steps: dynamic knowledge graph construction and knowledge graph augmented generation. 
	Initially, \ourapproach autonomously constructs a dynamic knowledge graph to store revised information while resolving potential knowledge conflicts. 
	Subsequently, it employs a fine-grained retrieval strategy coupled with an entity and relation detector to enhance  the accuracy of graph retrieval for LLM generation.
	Experimental results on benchmarks show that \ourapproach surpasses previous state-of-the-art models, delivering more accurate and reliable answers in environments with dynamic information.
\end{abstract}

%

\section{Introduction}
\label{sec:intro}
Large language models (LLMs) have gained widespread adoption due to their advanced language understanding and reasoning capabilities~\cite{survey1, survey2}.
However, as the world changes and information becomes outdated, the datasets or corpora used to pre-train LLMs may no longer be relevant, potentially leading to unreliable outcomes in certain question-and-answer applications. 
To circumvent the substantial time and costs associated with retraining LLMs, researchers have shifted focus to knowledge editing~\cite{KEsurvey1, KEsurvey2}. 
The goal of knowledge editing is to precisely modify and update the knowledge within LLMs in a cost-effective manner, thereby enabling them to deliver dependable answers.

\begin{figure}[t]
    \centering
    \includegraphics[width=\linewidth]{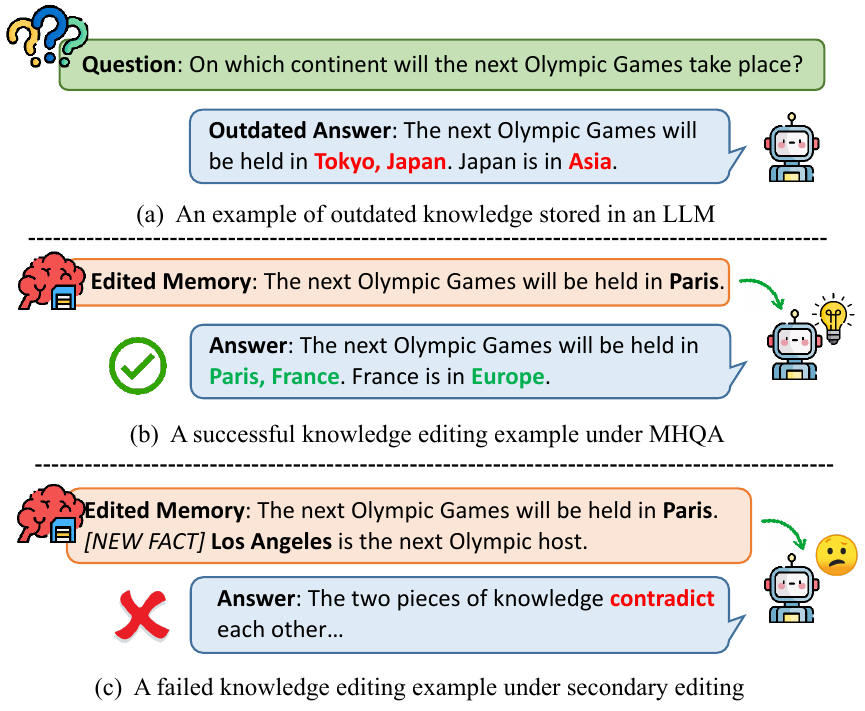}
    \caption{Examples of knowledge editing for MHQA.
    (a) an example of outdated information stored in LLMs.
    (b) a successful update with a parameter-preserving editing method.
    (c) a failure occurring during secondary editing.}
    \label{fig:intro}
\end{figure}

For LLMs, multi-hop question answering (MHQA) is a challenging task that requires high-order comprehension and reasoning abilities~\cite{MQAsurvey}. 
In particular, knowledge editing in MHQA is often vulnerable due to the \textbf{cascade effect}. 
For example, as shown in Figure~\ref{fig:intro}.(a) with the multi-hop question: \textit{``On which continent will the next Olympic Games take place?''}, 
if the information about the next Olympic host is updated from \textit{Tokyo} to \textit{Paris}, then the answer to the multi-hop question should also be updated to reflect \textit{Paris}' location.
Currently, approaches to knowledge editing for MHQA can be categorized into two main types:
(1) \textit{Parameter-based editing methods}:
These methods directly adjust the parameters of the LLM to change its output~\cite{ft2, rome, mend, memit}.
While effective for single-hop questions, recent studies have shown that they face challenges with MHQA, including issues like catastrophic forgetting and degradation of the original model's performance~\cite{hurt}.
(2) \textit{Parameter-preserving editing methods}:
These methods retain the original model's parameters and alter its behavior by adding extra modules~\cite{serac, t_patcher, mquake, pokemqa, temple_mqa}.
A typical strategy involves storing the edited fact directly in memory and using retrieval-enhanced techniques to prompt the LLM to modify its responses. Figure~\ref{fig:intro}.(b) illustrates a successful application of this approach in MHQA.

However, existing approaches exhibit two primary limitations. 
First, the retrieval process often relies on semantic similarity matching, which can be inadequate and overlook information pertinent to the original question. 
In multi-hop questions, only the initial entity is known, requiring the LLM to infer intermediate entities. 
Second, as illustrated in Figure~\ref{fig:intro}.(c), edited facts may require updates over time, a process we refer to as \textbf{secondary editing}. 
Directly storing secondarily edited knowledge in memory can conflict with previously edited facts, leading to interference in the retrieval process and errors during the LLM's inference phase.

To address the aforementioned limitations, we propose a novel method called \ourapproach:
\textbf{K}nowledge \textbf{E}diting with \textbf{D}ynamic \textbf{K}nowledge \textbf{G}raphs for MHQA.
Inspired by the evolving knowledge storage and convenient modification features of knowledge graphs, \ourapproach initially converts  edited knowledge into structured triples to establish a knowledge graph.
This knowledge graph can \textbf{dynamically} expand, modify, or delete information as the world's knowledge evolves.
Utilizing this dynamic knowledge graph, \ourapproach tackles the issue of secondary editing that may lead to potential conflicts, accurately providing the necessary edited knowledge for the LLM. 
When handling multi-hop questions, inspired by the chain-of-thought (CoT) approach~\cite{chainofthought}, \ourapproach use a fine-tuned LLM to decompose a multi-hop question into multiple sub-questions.
For each sub-question, \ourapproach perform fine-grained retrieval within the newly created knowledge graph and use detectors to filter the results, ensuring both the speed and accuracy of retrieval. 
Extensive experiments demonstrate that \ourapproach surpasses existing state-of-the-art knowledge editing methods.
We summarize the key contributions of our work as follows:
\begin{itemize}
    \item To the best of our knowledge, this is the first approach to automatically construct dynamic knowledge graphs specifically for knowledge editing, and to investigate the knowledge conflict issue of secondary editing.
    \item We propose a novel knowledge editing method,  named \ourapproach, for MHQA tasks that leverages knowledge graphs to resolve knowledge conflicts in secondary editing.
    \ourapproach employs a fine-grained retrieval and filtering strategy to improve the accuracy of the retrieval process.
    Additionally, the method is lightweight and applicable with all open-source and black-box LLMs.
    \item We conduct extensive experiments across various LLMs and datasets to validate the effectiveness and usability of \ourapproach. 
    Our empirical results and analysis demonstrate that \ourapproach significantly outperforms the advanced existing baselines, achieving superior performance.  
\end{itemize}

\begin{figure*}[t]
    \centering
    \includegraphics[width=\linewidth]{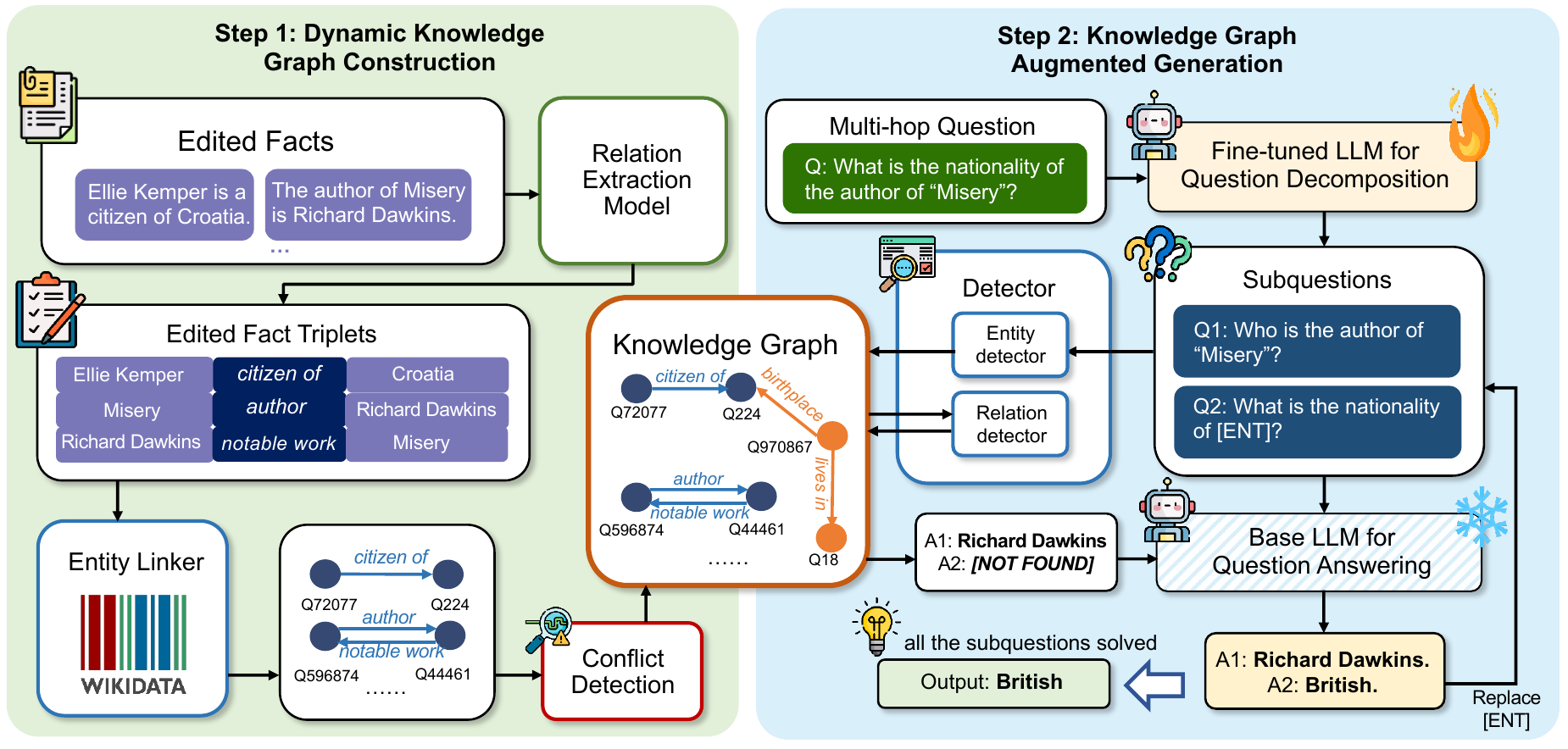}
    \caption{An overview of our proposed method \ourapproach, which consists of two main steps: Dynamic Knowledge Graph Construction and Knowledge Graph Augmented Generation.  
\ourapproach begins by dynamically building a new knowledge graph to store edited knowledge and address potential conflicts.
Subsequently, \ourapproach employs a fine-grained retrieval strategy, utilizing entity and relation detectors to enhance the accuracy of graph retrieval for LLM generation.}
    \label{fig:main}
\end{figure*}

\section{Related Work}

\subsection{Parameter-preserving editing methods}
Parameter-preserving editing methods do not modify the original model's parameters. 
Instead, they introduce additional modules (such as neurons, retrievers, etc.) to influence the model's behavior.
SERAC~\cite{serac} introduces a scope classifier to determine if an input request requires editing and employs a counterfactual model to handle requests.
T-patcher~\cite{t_patcher} and CaliNet~\cite{calinet} incorporate additional trainable parameters in feed-forward layers of PLMs to provide edited knowledge.
IKE~\cite{ike} utilizes in-context learning based on demonstration storage to prompt LLMs to edit knowledge with predefined templates.
MeLLo~\cite{mquake} is a simple editing method for multi-hop QA tasks, storing all edited facts externally while iteratively prompting LLMs to generate answers consistent with the edited facts.
PokeMQA~\cite{pokemqa} improves retrieval and model answer accuracy by dividing the task into 2 steps: question decomposition and decoupled self-checking.
TEMPLE-MQA~\cite{temple_mqa} directly generates multi-hop reasoning paths and constructs temporal graphs to enhance support for handling temporal multi-hop QA tasks.
Unlike existing solutions, \ourapproach constructs a structured knowledge graph to store the edited knowledge and supports dynamic updates to the edited knowledge.
\subsection{Parameter-based editing methods}
Parameter-based editing methods directly modify some or all parameters to incorporate edited knowledge into the model.
Current methods can be categorized into three types: fine-tuning, meta-learning, and locate-then-edit methods.
Traditional fine-tuning methods~\cite{ft1, ft2} can enable the model to learn new knowledge, but it suffers from catastrophic forgetting.
Meta-learning methods use a hyperparameter network to learn to modify model weights.
For example, MEND~\cite{mend} learns to transform the gradient of fine-tuned LLMs through low-rank decomposition of the gradient, but this approach incurs additional training overhead.
Locate-then-edit methods directly identify specific parameters within the model and modify them.
For instance, ROME~\cite{rome} uses causal mediation analysis to locate the editing region and modifies the feed-forward layer parameters in that region to update knowledge.
However, this method does not perform well for complex multi-hop QA problems.
Recent research~\cite{hurt} indicates that parameter-based editing methods are likely to significantly impact the model's original performance.

\section{Preliminaries}
\paragraph{Notations.}
We define each piece of factual knowledge embedded in the LLM as a triple $f=(s,r,o) \in F$, where $F$  represents the collection of all facts. 
Each triple $f$ consists of a subject $s$, a relation $r$ and an object $o$.
Edited knowledge shares the same subject and relation as the original knowledge, and the object is updated from $o$ to $o^*$, denoted as $e=(s,r,o \to o^*)$.
For example, a possible edit could be (Brazil, continent, South America $\to$ Asia).
A knowledge editing operation can be represented as $E=\{e_1,e_2,...,e_m\}$, potentially encompassing multiple edited triples of knowledge.

\paragraph{Multi-hop QA under knowledge editing.}
Given a multi-hop question $Q$, answering $Q$ requires sequentially querying and retrieving multiple facts.
Following the order in which these facts are retrieved, these facts can be represented a chain of facts $C=[(s_1,r_1,o_1),...,(s_n,r_n,o_n)]$, where $s_{i+1}=o_i$ and $o_n$ is the final answer to the question $Q$.
If we replace old knowledge in the chain of facts with new knowledge $(s_i,r_i,o_i^*)$, the cascading effect inherent in multi-hop question tasks affects the entire chain.
Consequently, the final reasoning chain becomes $C^*=[(s_1,r_1,o_1),...,(s_i,r_i,o_i^*),...,(s_n^*,r_n,o_n^*)]$, where $o_n^*$ is the updated answer.
It should be noted that any number of old knowledge triples can be replaced in the chain of facts.
The goal of the knowledge editing task for MHQA can be outlined as follows, given an editing set $E$ and a language model $M$, generate a conditionally edited language model $M^*$.
For each multi-hop question influenced by the editing operations in $E$, $M^*$ should be able to infer the correct answer $o_n^*$. 
Meanwhile, the reasoning chain needs to align with $C^*$, which we call \textbf{the golden path} for question $Q$.
\ourapproach freezes the parameters of $M$ and utilizes a dynamic knowledge graph $\mathcal{G}$ generated by the editing set $E$ as an external knowledge base to guide $M$ in generating the answer.

\section{Methodology}
\label{sec:method}
\ourapproach is a lightweight memory-based knowledge editing approach that stores edited knowledge in the form of a knowledge graph. 
This facilitates subsequent retrieval and enables dynamic updates to the graph  in response to real-world knowledge changes.
Additionally, we decouple multi-hop question decomposition from the original model to preserve its performance, handling this decomposition by fine-tuning an additional language model. 
As illustrated in the Figure~\ref{fig:main}, \ourapproach consists of  two main steps: (1) \textit{dynamic knowledge graph construction}, and (2) \textit{knowledge graph augmented generation}.
\subsection{Dynamic Knowledge Graph Construction}
\paragraph{Edited Knowledge Triples Extraction.}
Given a set of plain texts $T=\{t_1, t_2,...,t_n\}$ that represent edited knowledge in natural language form, \ourapproach employs a relation extraction model to transform these natural texts into structured triples.
This process can be described as follows:
\begin{equation}
    E_i=\{e_1^i,e_2^i,...,e_m^i\}=Retriever(t_i),i\in [1,n]
\end{equation}
where $Retriever$ is the relation extraction model.

Previous approaches typically stored natural texts directly in memory, retrieving the most relevant knowledge based on semantic similarity to the question. 
However, this method is often influenced by syntactic structures and a variety of comparable words, which significantly affects accuracy.
By pre-extracting relations from natural texts, we can more effectively capture the relationships between entities. 
As shown in Figure~\ref{fig:main}, from the sentence ``\textit{The author of Misery is Richard Dawkins}'',  we can extract two pieces of edited knowledge: (Misery, author, Richard Dawkins) and (Richard Dawkins, notable work, Misery), the latter of which is often overlooked by conventional editing methods.
\paragraph{Entity Linking.}
To store the edited knowledge in triple form into the knowledge graph, we note that different entity names might refer to the same entity (e.g., United States and U.S. denote the same entity). 
We link the extracted entities to Wikidata and store their corresponding node IDs and aliases in the knowledge graph.
\paragraph{Conflict Detection and Modification.}
Traditional knowledge editing methods do not account for the necessity of updating edited knowledge as the world evolves, a process we refer as \textbf{secondary editing}.
Secondary editing can introduce conflicting knowledge which, if stored directly in memory, negatively impacts subsequent retrieval and is difficult to detect.
The utilization of knowledge graphs can simplify this process.
For a new piece of edited knowledge $e_{new}=(s,r,o_{new}^*)$, we locate the old edited knowledge $e_{old}=(s,r,o_{old}^*)$ based on the subject $s$ and relation $r$, then remove it and add $e_{new}$ to the knowledge graph.
If $e_{old}$ is not found, this indicates that the knowledge has not undergone secondary editing, and $e_{new}$ can be added directly.
\subsection{Knowledge Graph Augmented Generation}
Unlike the MeLLo \cite{mquake} approach, which iteratively decomposes multi-hop questions, we propose an approach that utilizes a template $P_{divide}$ to guide a trained LLM to decompose a multi-hop question $Q$ into multiple sub-questions in a single step:
\begin{equation}
    \{q_1,q_2,...,q_n\} = LLM(P_{divide}(Q))
\end{equation}
where $P_{divide}(Q)$ is the prompt obtained by filling the multi-hop question $Q$ into the template $P_{divide}$.
Except for the first sub-question $q_1$, which contains the subject $s$ from $Q$, the subjects in other sub-questions are replaced by a special marker $[ENT]$.
We believe this can effectively reduce the burden on LLMs in understanding contexts during iteration, thereby improving the accuracy of question decomposition.

To search for relevant edited knowledge in the knowledge graph, we employ a fine-grained retrieval scheme and filter the retrieval results using \textbf{entity and relation detectors}.
When processing each generated sub-question $q_i$, we first use a relation extraction model to generate a set of possible entities $K=\{k_1,...k_n\}$.
Then, we use an entity detector $g_\phi(q_i,k_j)\to [0,1]$ to predict the probability that entity $k_j$ is the subject of $q_i$.
We select $s=\arg\max\limits_{k_i\in K}{g_\phi(q_i,k_j)}$ as the subject and link it to the knowledge graph.
If the subject does not exist in the knowledge graph, it indicates that no information related to the subject needs to be modified, and we directly call the original LLM for output.
Otherwise, we can easily obtain the set of relations $R=\{r_1,...,r_m\}$ related to the entity and use a relation detector $g_\psi(q_i,r_j)\to [0,1]$ to detect the probability of the relation $r_j$ being relevant to the subject in $q$.
If the highest probability $p$ exceeds a threshold $\alpha$, which is set to 0.5 in our experiments,  we can retrieve the corresponding fact triple $(s,r,o^*)$ and use $o^*$ as the retrieval answer.
Otherwise, we directly call the original LLM for the answer.
The final answer $a_i$ to the sub-question $q_i$ can be expressed using the formula:
\begin{equation}
    a_i=
    \begin{cases}
        LLM(P_{retrieve}(q_i,o^*)), & \text{if } p > \alpha \\
        LLM(P_{answer}(q_i)), & \text{if} \ s \notin G \ \text{or} \ p <= \alpha
    \end{cases}
\end{equation}
\begin{equation}
    o^*= \mathcal{G}(s, \arg\max\limits_{r_j\in R}{g_\psi(q_i, r_j)})
\end{equation}
where $P_{answer}$ is the template guiding the original LLM in generating answers, $P_{retrieve}$ is a template that prompts the LLM to refine the response based on the retrieved entities, and $\mathcal{G}$ is the knowledge graph constructed in Step 1.

Entity detector and relation detector can be implemented using any text classification model.
After obtaining the answer $a_i$ to $q_i$, we use $a_i$ as the subject to fill in the special marker in the next sub-question $q_{i+1}$.
This process is iteratively repeated until all sub-questions are resolved, culminating in the output of the final answer.

\subsection{Training Objectives}
\paragraph{Question Decomposition Module.}

This module is designed to enable the LLM to learn both explicit and implicit atomic sentence meanings and the relationships between sentences. 
It enhances the LLM's ability to decompose a complex question into multiple sub-questions.
This objective focuses on training the LLM to predict subsqequent tokens based on previous tokens.
Specifically, given the predefined prompt template $\bm{p}$ and question $\bm{q}$, the objective function for generating the sub-questions sequence $o=[o_1,...,o_T]$ is outlined as follows:

\begin{equation}
	\mathcal{L}_\text{dec} (\theta) =  - \sum_{t=1}^{T}\log p_{\theta} ({o}_{t}|\bm{o}_{<t}, \bm{p}, \bm{q})
\end{equation}
where $\theta$ represents the parameters of the model, and the output sequence $o=[q_1,...,q_H]$ is composed of multiple sub-clauses derived from decomposition.
$q_i$ represents each individual sub-question,  and $H$ represents the number of sub-questions.

\paragraph{Entity detector and relation detector.}
Specifically, given the input pair $(q_i,k_j)$ or $(q_i,r_j)$, the training objective of the entity detector $g_\phi$ and relation detector $g_\psi$ is to minimize the average binary cross-entropy loss over the training dataset:
\begin{equation}
    \mathcal{L}_\text{det}=-\frac{1}{N}\sum_{i=1}^{N}[y_{i}\log(p_i)+(1-y_i)\log(1-p_i)]
\end{equation}
where $N$ denotes the number of training samples. $y_i$ represents the true label of the $i$-th sample, and $p_i$ represents the probability that the $i$-th sample is predicted to be a subject or relation in sub-question $q_i$.
For more details about training, please refer to Appendix C and D.

\begin{table*}[t]
	\centering
		\begin{tabular}{l|cccccc|cccc}
			\toprule
			\multicolumn{1}{c|}{} &
			\multicolumn{6}{c|}{MQUAKE-CF-3K} &
			\multicolumn{4}{c}{MQUAKE-T} \\ \cline{2-7} \cline{8-11} 
			\multicolumn{1}{c|}{\textbf{Method}}  &
			\multicolumn{2}{c|}{1 edited} &
			\multicolumn{2}{c|}{100 edited} &
			\multicolumn{2}{c|}{All edited} &
			\multicolumn{2}{c|}{1 edited} &
			\multicolumn{2}{c}{All edited} \\ \cline{2-7} \cline{8-11} 
			\multicolumn{1}{c|}{} &
			M-Acc. &
			H-Acc. &
			M-Acc. &
			H-Acc. &
			M-Acc. &
			H-Acc. &
			M-Acc. &
			H-Acc. &
			M-Acc. &
			H-Acc. \\ \hline
			\multicolumn{11}{c}{\textbf{LLaMa 2-7B}}  \\ \hline
			FT$_{\mathrm{COT}}\ast$ &
			22.30 &
			- &
			2.13 &
			- &
			OOM &
			- &
			47.32 &
			- &
			3.75 &
			- \\
			FT$\ast$ &
			28.20 &
			7.30 &
			2.37 &
			0.03 &
			OOM &
			OOM &
			56.48 &
			33.89 &
			1.02 &
			0.37 \\
			ROME$_{\mathrm{COT}}\ast$ &
			11.17 &
			- &
			2.87 &
			- &
			2.77 &
			- &
			28.96 &
			- &
			14.40 &
			- \\
			ROME$\ast$ &
			13.13 &
			5.37 &
			3.50 &
			0.03 &
			3.63 &
			0.10 &
			24.89 &
			17.99 &
			1.71 &
			0.32 \\
			MEMIT$_{\mathrm{COT}\ast}$ &
			11.83 &
			- &
			9.23 &
			- &
			5.57 &
			- &
			36.88 &
			- &
			31.58 &
			- \\
			MEMIT$\ast$ &
			14.97 &
			6.43 &
			9.40 &
			2.47 &
			2.30 &
			0.37 &
			30.89 &
			23.98 &
			25.21 &
			20.13 \\
			MeLLo$\dag$ &
			33.57 &
			9.90 &
			20.00 &
			10.07 &
			17.33 &
			9.90 &
			65.78 &
			55.27 &
			57.69 &
			44.55 \\ 
			PokeMQA$\ast$  &
			\underline{44.13} &
			\underline{30.60} &
			\underline{37.33} &
			\underline{27.83} &
			\underline{32.83} &
			\underline{23.87} &
			\textbf{75.43} &
			\underline{60.44} &
			\textbf{74.36} &
			\underline{60.22} \\ 
			\ourapproach (Ours) &
			\textbf{66.80} & \textbf{63.67} & \textbf{58.50} & \textbf{55.37} & \textbf{48.30} & \textbf{43.90} &
			\underline{73.13} & \textbf{69.06} & \underline{71.15} & \textbf{66.76}\\
			\hline
			\multicolumn{11}{c}{\textbf{Vicuna-7B}}  \\ \hline
			MeLLo$\ast$ &
			30.70 &
			20.84 &
			24.75 &
			12.25 &
			22.35 &
			10.18 &
			60.72 &
			48.55 &
			51.55 &
			42.97 \\
			PokeMQA$\ast$ &
			\underline{45.83} &
			\underline{34.80} &
			\underline{38.77} &
			\underline{31.23} &
			\underline{31.63} &
			\underline{25.30} &
			\textbf{74.57} &
			\underline{55.19} &
			\underline{73.07} &
			\underline{55.09} \\  
			\ourapproach (Ours) &
			\textbf{68.60} & \textbf{65.13} & \textbf{62.43} & \textbf{58.20} & \textbf{51.10} & \textbf{44.67} &
			\underline{71.90} & \textbf{67.23} & \textbf{74.68} & \textbf{66.54}
			\\
			\hline
			\multicolumn{11}{c}{\textbf{GPT-3.5-turbo-instruct}}\\ \hline
			MeLLo$\ast$ &
			57.43 &
			28.80 &
			40.87 &
			28.13 &
			35.27 &
			25.30 &
			\textbf{88.12}  &
			52.84 &
			74.57 &
			53.53 \\
			PokeMQA$\ast$ &
			\underline{67.27} &
			\underline{56.37} &
			\underline{56.00} &
			\underline{49.63} &
			\underline{45.87} &
			\underline{39.77} &
			76.98 &
			\underline{68.09} &
			\textbf{78.16} &
			\underline{67.88} \\ 
			\ourapproach (Ours) &
			\textbf{68.00} & \textbf{65.33} & \textbf{59.50} & \textbf{56.80} & \textbf{49.10} & \textbf{43.17} &
			\underline{78.75} & \textbf{76.18} & \underline{77.19} & \textbf{73.77} 
			\\
			\bottomrule
		\end{tabular}

	\caption{The experimental results on MQUAKE-CF-3K and MQUAKE-T benchmarks.
		The best results are indicated in \textbf{bold}, and the second-best results are \underline{underlined}.
		Results marked with $\ast$ indicate they are sourced from~\cite{pokemqa}, and results marked with $\dag$ indicate they are sourced from~\cite{temple_mqa}.
		The notation ``$k$ edited'' refers to the size of the edit batch is 
		$k$.
		``COT'' implies that the current method employs chain-of-thought prompting, otherwise, a decomposition prompt is used.
		The evaluation metrics are multi-hop accuracy (M-Acc.) and hop-wise answering accuracy (H-Acc.).
		``OOM'' denotes that the method does not work due to high GPU memory usage.
	}
	
	\label{main experiment}
\end{table*}

\section{Experiments}
\subsection{Experimental Settings}
\paragraph{Datasets.}
We evaluate \ourapproach using the MQuAKE dataset.
MQuAKE is a knowledge editing benchmark for multi-hop QA, comprising MQuAKE-CF based on counterfactual editing and MQuAKE-T based on temporal knowledge updates.
We use MQuAKE-CF as the training set, which contains 9,218 data points, and MQuAKE-CF-3k as the test set, which includes 3,000 data points.
It is important to note that there are no overlapping data points between these two sets.
These datasets include numerous $k$-hop questions ($k\in \{2,3,4\}$), with each question corresponding to $n$ edits ($n\in[1,k]$).
See Appendix A for more details.
\paragraph{Baselines.}
We compare \ourapproach with common knowledge editing methods, including parameter-preserving and parameter-based editing methods.
The parameter-preserving editing methods include MeLLo~\cite{mquake} and PokeMQA~\cite{pokemqa}, while the parameter-based editing methods include FT~\cite{ft2}, ROME~\cite{rome}, and MEMIT~\cite{memit}.
We also report the performance of these methods under chain-of-thought (CoT) and question decomposition (QD) prompts.
See Appendix B for more details.
\paragraph{Evaluation Metrics.}
Following previous work~\cite{mquake,pokemqa}, we use Multi-hop Accuracy (M-Acc) and Hop-wise Answering Accuracy (H-Acc) as evaluation metrics.
Multi-hop accuracy measures the accuracy of the LLM in answering multi-hop questions.
However, in some cases, the LLM might produce an incorrect reasoning process but still arrive at the correct answer.
Hop-wise answering accuracy measures the accuracy where every answer in the reasoning path is correct, avoiding such interference.
This is the primary metric we focus on to assess the model's ability to use edited knowledge.
For all metrics, higher values indicate better performance.

\begin{table*}[t]
	\centering
		\begin{tabular}{ccc|cccccc|cccc}
			\toprule
			\multicolumn{3}{l|}{} &
			\multicolumn{6}{c|}{MQUAKE-CF-3k} &
			\multicolumn{4}{c}{MQUAKE-T}
			\\ \cline{4-13} 
			\multicolumn{1}{c}{\textbf{$g_{\phi}$}} &
			\multicolumn{1}{c}{\textbf{$g_{\psi}$}} &
			\multicolumn{1}{c|}{\textbf{CDM}} &
			\multicolumn{2}{c|}{1 edited} &
			\multicolumn{2}{c|}{100 edited} & \multicolumn{2}{c|}{All edited} &
			\multicolumn{2}{c|}{1 edited} &
			\multicolumn{2}{c}{All edited}
			\\ \cline{4-13} 
			&&&
			M-Acc. &
			H-Acc. &
			M-Acc. &
			H-Acc. &
			M-Acc. &
			H-Acc. &
			M-Acc. &
			H-Acc. &
			M-Acc. &
			H-Acc. 
			\\ \hline
			- &$\sqrt{}$&$\sqrt{}$&
			35.23 &
			31.63 &
			31.17 &
			26.93 &
			23.10 &
			18.87 &
			54.39 &
			50.05 &
			53.80 &
			48.50
			\\
			$\sqrt{}$ &-&$\sqrt{}$&
			20.03 &
			16.03 &
			19.13 &
			15.10 &
			16.80 &
			12.57 &
			54.98 &
			52.19 &
			56.48 &
			52.78
			\\
			$\sqrt{}$ & $\sqrt{}$& -&
			66.30 &
			63.43 &
			58.43 &
			54.60 &
			41.10 &
			36.83 &
			73.13 &
			68.74 &
			71.15 &
			63.81
			\\ \hline
			$\sqrt{}$ &$\sqrt{}$&$\sqrt{}$&
			\textbf{66.80} &
			\textbf{63.67} &
			\textbf{58.50} &
			\textbf{55.37} &
			\textbf{48.30} &
			\textbf{43.90} &
			\textbf{73.13} &
			\textbf{69.06} &
			\textbf{71.15} &
			\textbf{66.76}
			\\
			\bottomrule
		\end{tabular}
	
	\caption{Ablation results of \ourapproach and its variants in terms of Multi-hop Accuracy (M-Acc.) and Hop-wise Answering Accuracy (H-Acc.) on MQUAKE datasets. ``CDM'' refers to the Conflict Detection and Modification module.}
	\label{tab:ablation}
\end{table*}

\paragraph{Experimental setup.}
We train an entity detector and a relation detector based on the DistilBERT~\cite{distilbert} model and fine-tune the Llama 2-7B model for the question decomposition task.
In addition, we use REBEL~\cite{rebel} as our relation extraction model and \textit{spacy entity linker} as entity linking model.
It is important to note that there is no overlap between the training and test datasets we use.
We conducted training on multiple base models, including mainstream open-source models Llama 2-7B~\cite{llama2} and Vicuna-7B~\cite{vicuna}, as well as the black-box model GPT-3.5-turbo-instruct.
Parameter-preserving editing methods are applicable to all models, while parameter-based editing methods are only applicable to open-source models.
Therefore, we only reported the experimental results of parameter-based editing methods on open-source models.
All our experiments are carried out on a NVIDIA 8$\times$A800-SXM4-80G machine.
\subsection{Main Results}
Table~\ref{main experiment} shows the experimental results on MQUAKE datasets.
We can draw the following observations:

(1) \ourapproach demonstrates outstanding performance across all metrics, notably surpassing all baselines in the H-Acc metric.
Utilizing Llama 2-7B as the base model, 
\ourapproach achieves an improvement of 108.1\% in 1 edit batch, 96.2\% in 100 edit batches, and 83.9\% in all edit batches compared to the best previous baseline. 
This robust performance  underscores \ourapproach's capability to effectively answer multi-hop questions under conditions of knowledge editing and adapt seamlessly to varying numbers of edits.

(2) Notably, other methods exhibit a significant discrepancy between M-Acc and H-Acc, indicating that their reasoning processes do not align with the task logic presented in the prompt, which can be seen as a mismatch in question decomposition.
In contrast, \ourapproach demonstrates a smaller difference between M-Acc and H-Acc, primarily because we fine-tune the model specifically on the question decomposition task, decoupling it from the original model and reducing the burden of understanding complex prompts.
This results in more reliable reasoning outcomes.
  
 (3) Interestingly, on the MQUAKE-CF-3k benchmark, \ourapproach's performance with a 7B parameter open-source base model matches or even surpasses that of GPT-3.5-turbo-instruct.
This comparison reveals that the knowledge capability of the 7B parameter model is on par with that of the black-box model, though the latter demonstrates a stronger ability to understand and follow human instructions. 
\ourapproach effectively bridges this gap, enhancing the 7B parameter model's ability to adeptly handle multi-hop QA problems under conditions of knowledge editing.

(4) We also observe that all baselines perform best under the 1 edit batch setting, with performance declining as the edit batch size increases,  a trend that is particularly pronounced in the MQUAKE-CF-3k dataset.
This phenomenon can be attributed to two main factors: 
first, conducting retrieval within large batches of edits is a challenging task, especially when the entities and relationships involved are very similar.
Second, in large batch edits, the edited knowledge in different cases within the MQUAKE-CF-3k dataset may conflict with each other.
Thanks to the integration of a conflict detector and the application of dynamic knowledge graphs, 
we have effectively improved \ourapproach's performance in scenarios involving large batch edits. 
Even under the all edit batch setting, \ourapproach can still achieve an accuracy rate close to 50\%.

\begin{table*}[ht]
	\centering
	\small
	\noindent\fbox{%
		\begin{minipage}{2.0\columnwidth}
			\ttfamily
			
			Edited Knowledge: \textbf{Association football was created in the country of Brazil.
				Brazil is located in the continent of Africa.} \\
			Knowledge Graph: (Association football, country of origin, Brazil); (Brazil, sport, Association football); (Brazil, continent, Africa) 
			\\
			\\
			Question: \textbf{\underline{What is the continent of origin for the sport associated with Watford F.C.?}}\\
			Subquestions: \sethlcolor{lightyellow}\hl{1. Which sport is Watford F.C. associated with? 2. Which country was [ENT] created in? 3. Which continent is [ENT] located in?}\\
			Thought 1: Which sport is Watford F.C. associated with?\\
			Retrieved answer: \sethlcolor{lightblue}\hl{[None]} \\
			Generated answer: \sethlcolor{lightgreen}\hl{Watford F.C. is associated with Association Football (Soccer).}\\
			Thought 2: Which country was Association Football (Soccer) created in?\\
			Retrieved answer: \sethlcolor{lightblue}\hl{Brazil.} \\
			Generated answer: \sethlcolor{lightgreen}\hl{Association Football (Soccer) was created in Brazil.}\\
			Thought 3: Which continent is Brazil located in?\\
			Retrieved answer: \sethlcolor{lightblue}\hl{Africa.} \\
			Generated answer: \sethlcolor{lightgreen}\hl{Brazil is located in Africa.}\\
			Final answer: \textbf{\underline{Africa}}
		\end{minipage}
	}
	
	\caption{A case study of \ourapproach solving one 3-hop question in MQUAKE-CF-3K.
		Yellow parts are generated by the fine-tuned LLM for question decomposition.
		Green parts are generated by the base LLM for question answering, and blue parts are answers retrieved in the knowledge graph.}
	\label{tab:case}
\end{table*}

\begin{figure}[t]
	\centering
	\includegraphics[width=\linewidth]{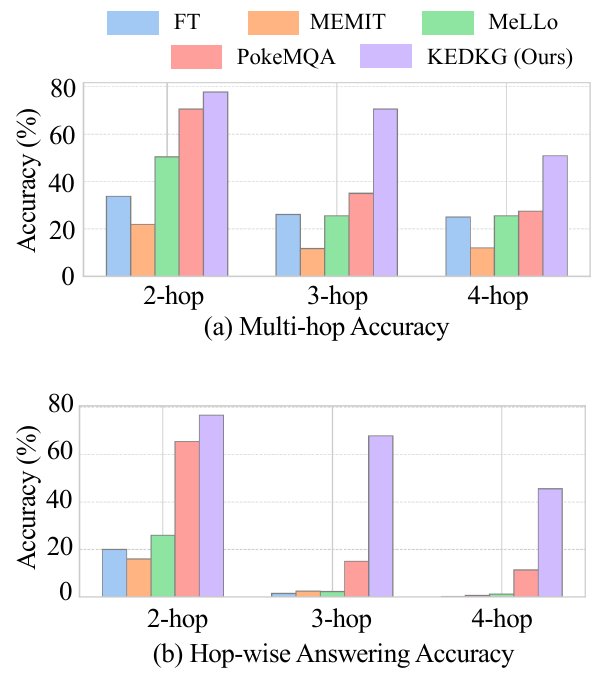}
	\caption{Multi-hop Accuracy and Hop-wise Answering Accuracy results on MQUAKE-CF-3K, utilizing different knowledge editing methods. The experiments are conducted on Llama 2-7B and the edit batch size is 1.}
	\label{fig:ablation}
\end{figure}

\subsection{Ablation Study}
We conduct ablation experiments with the base model Llama 2-7B on the MQUAKE-CF-3k and MQUAKE-T datasets to analyze the impact of $g_{\phi}$, $g_{\psi}$, and conflict detection modification (CDM) modules.

\paragraph{The trained entity detector $g_{\phi}$ and relation detector $g_{\psi}$ have proven to be effective.}
As shown in Table~\ref{tab:ablation}, replacing $g_{\phi}$ or $g_{\phi}$ with the semantic similarity model, all-MiniLM-L6-v2 from Huggingface~\cite{huggingface}, resulted in a decrease in answer accuracy across all experimental settings.
The most significant performance decline was observed when the relation detector was replaced, with a 31.5\% drop in Multi-hop Accuracy under the ``all edited" setting. 
This decline highlights the limitations of relying solely on semantic similarity to select relations from the knowledge graph, as a specific relation may have various expressions, making such an approach unreliable.

\paragraph{The CDM module effectively enhances performance in batch editing.}
The MQUAKE-CF dataset includes some conflicting edits, providing an opportunity to evaluate \ourapproach's conflict resolution capabilities. 
As shown in Table~\ref{tab:ablation}, the CDM module has no significant effect in small batch editing, which is intuitive since there are rarely conflicting edits in small batches.
However, as the number of edits increases, the probability of conflicts also rises.
Removing the CDM module led to a 7.2\% drop in Multi-hop Accuracy.
This decrease is attributed to the presence of conflicting knowledge triples within the knowledge graph, which disrupt the retrieval process.
Introducing the CDM module effectively minimizes this interference and facilitates the execution of secondary edits.
On the MQUAKE-T dataset, we observe that applying the CDM module does not significantly impact the results. This is intuitive because MQUAKE-T itself does not contain conflicting edits.

\subsection{Analysis}
\paragraph{\ourapproach excels in 3-hop and 4-hop question answering tasks.}
As shown in Figure~\ref{fig:ablation}, both \ourapproach and PokeMQA perform well in two-hop questions.
However, in 3-hop and 4-hop questions, \ourapproach significantly outperforms PokeMQA,  almost doubling the performance metrics. 
In the most complex 4-hop questions, \ourapproach achieves over 50\% Multi-hop Accuracy, while the accuracy of other methods remains below 25\%.
This superior performance is attributed to our training on the decomposition task, which effectively minimizes the likelihood of errors or hallucinations in the LLM during the decomposition process. 
Additionally, the fine-grained retrieval based on the knowledge graph significantly improves the accuracy of answers at each hop.

Meanwhile, we have also observed that \textbf{parameter-based editing methods generally underperform compared to parameter-preserving editing methods in multi-hop QA tasks}, exhibiting notably low Hop-wise Answering Accuracy. 
This suggests that current parameter-based editing methods still struggle to enable LLMs to flexibly utilize edited knowledge during the reasoning process.

\subsection{Case Study}
We conduct a case study as presented in Table~\ref{tab:case}.
\ourapproach first extracts edited triples from the given edited knowledge.
After passing through the entity linking and conflict detection modules, these triples are stored in the knowledge graph $\mathcal{G}$.
Note that Table~\ref{tab:case} only shows the edited knowledge relevant to this case.
In large-scale editing, there are many instances of edited knowledge and conflicting edits.

In the knowledge graph-based QA phase, the fine-tuned LLM used for question decomposition breaks down the multi-hop question into three sub-questions, where the subjects of the latter two sub-questions are replaced with special markers.
Subsequently, the entity detector and relation detector are applied sequentially to retrieve answers for the three sub-questions.
For the first sub-question, since no relevant facts about \textit{Watford F.C.} are found in $\mathcal{G}$, the question is directly handed to the base LLM to produce an answer, and the entity in the answer is then filled into the next sub-question.
For the latter two questions, \ourapproach successfully retrieves the edited facts from $\mathcal{G}$ and prompts the LLM to provide answers, ultimately yielding the correct answer.

\section{Conclusion}
In this paper, we propose a novel knowledge editing method, \ourapproach for MHQA.
\ourapproach constructs a dynamic knowledge graph to store edited knowledge and to handle conflicting edits.
Additionally, \ourapproach employs a fine-grained retrieval scheme to fetch edited knowledge from the knowledge graph, guiding the model to modify its answers.
Extensive experiments on the MQUAKE dataset demonstrate that \ourapproach outperforms previous knowledge editing methods.
\section*{Acknowledgements}
This work was supported in part by National Science Foundation of China (62476070), Shenzhen Science and Technology Program (JCYJ20241202123503005, GXWD20231128103232001) and Department of Science and Technology of Guangdong (2024A1515011540). This work was also supported in part by the Major Key Project of PCL under Grant PCL2024A06 and PCL2022A05, and in part by the Shenzhen Science and Technology Program under Grant RCJC20231211085918010. This research is supported by the NSFC (Grant: 62376074), the Shenzhen Science and Technology Program~(Grants: SGDX20230116091244004, JSGGKQTD20221101115655027, KJZD20231023095959002).

\bibliography{aaai25}

\appendix
\section{Datasets}
Table~\ref{datasetstatistic} provides an overview of the statistics for the MQUAKE-CF and MQUAKE-T datasets.
The MQUAKE-CF dataset contains over 9,000 $N$-hop questions (with $N$ being 2, 3, or 4), each linked to one or more edits.
The MQUAKE-T dataset comprises 1,800 instances, each corresponding to a change in a real-world fact.

\begin{table}[h]
\resizebox{0.98\linewidth}{!}{
    \footnotesize
\begin{tabular}{llcccc}
\toprule
 \textbf{Datasets} & \textbf{\#Edits} & \multicolumn{1}{l}{\textbf{2-hop}} & \multicolumn{1}{l}{\textbf{3-hop}} & \multicolumn{1}{l}{\textbf{4-hop}} & \multicolumn{1}{l}{\textbf{Total}} \\ \hline
                                 & 1       & 513  & 356  & 224  & 1093 \\
                                 & 2       & 487  & 334  & 246  & 1067 \\
\multicolumn{1}{c}{MQUAKE-CF-3K} & 3       & -    & 310  & 262  & 572  \\
                                 & 4       & -    & -    & 268  & 268  \\
                                 & All     & 1000 & 1000 & 1000 & 3000 \\ \hline
\multicolumn{1}{c}{MQUAKE-T}     & 1 (All) & 1421 & 445  & 2    & 1868 \\
\bottomrule
\end{tabular}
}
\caption{Statistics of MQUAKE datasets used in experiments.}
\label{datasetstatistic}
\end{table}
\section{Baselines}
We compare \ourapproach with two types of baselines: 1) \textit{Parameter-preserving editing methods} and 2) \textit{Parameter-based editing methods}.
The details of each baseline are described below:
\subsection{Parameter-preserving editing methods}
\begin{itemize}
    \item MeLLo~\cite{mquake} employs the plan-and-solve approach.
    For each sub-problem, MeLLo initially prompts the model to produce a potential answer and then provides the model with the retrieved edits to assess their relevance.
    \item PokeMQA~\cite{pokemqa} extends MeLLo. It prompts LLMs to decompose knowledge-augmented multi-hop question, while interacting with a detached trainable scope detector to modulate LLMs behavior depending on external conflict signal.
\end{itemize}
\subsection{Parameter-based editing methods}
\begin{itemize}
    \item FT~\cite{ft2} involves applying a gradient descent algorithm to the model using the new knowledge in order to update its parameters.
    \item ROME~\cite{rome} first identifies the factual knowledge in a particular layer of the transformer architecture, and then adjusts the feed-forward network in that layer to embed the new information.
    \item MEMIT~\cite{memit} extends ROME by adjusting multiple layers of feedforward networks, allowing it to modify a substantial amount of knowledge.
\end{itemize}
\section{Training Dataset Construction}
We use the MQUAKE-CF dataset as the source for our training set, filtering out cases that share the same $(s, r)$ pairs with MQUAKE-CF-3k and MQUAKE-T to ensure that there is no overlap between the training and test sets.

To train the entity detector, we perform entity extraction for each sub-question $q_i$ in the dataset, resulting in a set of potential entities $K=\{k_1,...,k_n\}$.
Each entity in this set was then compared to the true subject $s$ in $q_i$, creating the training set for the entity detector $D_e=\{(q_1,k_1,v_{1,1}),...,(q_i,k_j,v_{i,j})\}$, where $v$ can take values of either 0 or 1, indicating whether the entity is the true subject of $q_i$.

For training the relation detector, we identify all the relationships $R=\{r_1,...,r_m\}$ associated with the target subject $s$ in the constructed knowledge graph $\mathcal{G}$ and the corresponding objects $O^*=\{o_1^*,...,o_m^*\}$.
We then sequentially check if each object matches the target object, constructing the training set for the relation detector $D_r=\{(q_1,r_1,v_{1,1}),...,(q_i,r_j,v_{i,j})\}$, where $v$ indicates whether the relationship exists in $q_i$.

Additionally, the MQUAKE-CF dataset provides three multi-hop questions $Q$ and their corresponding sub-questions $q$ for each case.
We retained the first sub-question of each multi-hop question and replaced the subject $s$ in the remaining sub-questions with a special marker.
This process allowed us to construct a dataset $D_L=\{(Q_1,q_1),...,(Q_n,q_n)\}$ for training the LLM to handle question decomposition tasks.
\section{Details about Training}

Table~\ref{implement} reports the training parameters for the LLM and the entity and relation detectors.
We split the MQUAKE-CF dataset into 80\% for the training set and 20\% for the validation set to evaluate the effectiveness of the model training.
The results show that the performance of the entity and relation detectors exceed 99\%.

\begin{table}[ht]
\centering
\small
\renewcommand{\arraystretch}{1}

\resizebox{\columnwidth}{!}{
\begin{tabular}{cccc}
\toprule
\multicolumn{1}{c}{\textbf{Hyper-parameters}}& \textbf{LLM} & \textbf{$g_{\phi}$} & \textbf{$g_{\psi}$} \\ \hline
\multicolumn{1}{c}{epoch}  & 3  & 3  & 3 \\
\multicolumn{1}{c}{sequence length}  & 2048  & -  & - \\
\multicolumn{1}{c}{learning rate}  & 1e-5  & 2e-5  & 2e-5 \\
\multicolumn{1}{c}{batch size}  & 1  & 16  & 16 \\
\multicolumn{1}{c}{optimizer}  & AdamW  & AdamW  & AdamW \\
\multicolumn{1}{c}{weight decay}  & 0.01  & 0.01  & 0.01 \\
\bottomrule

\end{tabular}%
}
\caption{
Hyper-parameters of training.
\textbf{LLM} refers to the fine-tuned LLM for question decomposition.
$g_{\phi}$ refers to the entity detector and $g_{\psi}$ refers to the relation detector.}

\label{implement}
\end{table}

\begin{table*}[ht]
    \centering
    \small
    \noindent\fbox{%
    \begin{minipage}{2.0\columnwidth}
    \ttfamily
    
You need to divide a multi-hop question into several sub-questions.\\
Examples) \\
Question: Who is the head of state of the country where Rainn Wilson holds a citizenship? \\
Subquestion: \\
What is the country of citizenship of Rainn Wilson?\\
What is the name of the current head of state in [ENT]?\\

Question: What is the capital city of the country of citizenship of Ivanka Trump's spouse?\\
Subquestion: \\
Who is Ivanka Trump's spouse?\\
What is the country of citizenship of [ENT]?\\
What is the capital city of [ENT]?\\
        \sethlcolor{lightgray}\hl{$[\text{4 } $in-context demonstrations abbreviated$]$} \\
        Question: <<<<QUESTION>>>>\\
        Subquestion:
    \end{minipage}
    }

    \caption{Prompts for question decomposition.}
    \label{p1}
\end{table*}

\begin{table*}[ht]
    \centering
    \small
    \noindent\fbox{%
    \begin{minipage}{2.0\columnwidth}
    \ttfamily
    
For each question, provide a short and accurate answer.\\
Examples) \\
Question: Which country was jazz created in? \\
Answer: United States of America.\\

Question: What is the capital of France?\\
Answer: Paris.\\

Question: What is the largest planet in our solar system?\\
Answer: Jupiter.\\
        \sethlcolor{lightgray}\hl{$[\text{6 } $in-context demonstrations abbreviated$]$} \\
        Now, answer the following question in the same format:\\
        Question: <<<<QUESTION>>>>
    \end{minipage}
    }

    \caption{Prompts for sub-question answering.}
    \label{p2}
\end{table*}
\begin{table}[ht]
\centering
\small
\renewcommand{\arraystretch}{1.3}

\resizebox{\columnwidth}{!}{
\begin{tabular}{cccccc}
\toprule
\multicolumn{1}{c}{\textbf{Hyper-parameter}}& \textbf{0.1} & \textbf{0.25} & \textbf{0.5} & \textbf{0.75} & \textbf{0.9} \\ \hline
\multicolumn{1}{c}{$\alpha$}  & 64.24  & 66.13 & \textbf{66.80} & 66.60 & 64.80 \\
\bottomrule

\end{tabular}%
}
\caption{
The effect of the hyper-parameter $\alpha$ on the results, with the base model being Llama 2-7B and using a single edit batch setting.}

\label{hyper}
\end{table}

\section{Prompts for \ourapproach}
Table~\ref{p1} and~\ref{p2} provide the prompts used for multi-hop question decomposition and sub-question answering with the LLM, respectively.
When we successfully retrieve the edited knowledge from the constructed knowledge graph, we directly provide it to the LLM as the retrieved fact.

\section{Hyper-parameter Analysis}

We conduct a hyper-parameter analysis of the threshold $\alpha$ in Section 4.2.
The results in Table~\ref{hyper} indicate that the best performance is achieved when alpha is set to 0.5.
This suggests that a smaller alpha value may lead to the retrieval of incorrect edited knowledge, while a larger alpha value might filter out some useful information.

\end{document}